\documentclass{vgtc}               

\ifpdf
  \pdfoutput=1\relax                   
  \usepackage{graphicx}                
  \DeclareGraphicsExtensions{.pdf,.png,.jpg,.jpeg} 
\else
  \usepackage{graphicx}                
  \DeclareGraphicsExtensions{.eps}     
\fi%

\graphicspath{{figures/}{pictures/}{images/}{./}} 
\usepackage{subfig}
\usepackage{microtype}                 
\PassOptionsToPackage{warn}{textcomp}  
\usepackage{textcomp}                  
\usepackage{mathptmx}                  
\usepackage{times}                     
\usepackage{cite}                      
\usepackage{tabu}                      
\usepackage{booktabs}                  
\usepackage{amssymb}

\usepackage{tikz}

\onlineid{1008}

\vgtccategory{Research}

\vgtcinsertpkg

\title{DM-VTON: Distilled Mobile Real-time Virtual Try-On}

\author{
Khoi-Nguyen Nguyen-Ngoc\thanks{e-mail: 19120106@student.hcmus.edu.vn}\\%
    \parbox{1.4in}{\scriptsize \centering University of Science \\ VNU-HCM, Vietnam} \\%
\and 
Thanh-Tung Phan-Nguyen\thanks{e-mail: 19120424@student.hcmus.edu.vn}\\%
    \parbox{1.4in}{\scriptsize \centering University of Science \\ VNU-HCM, Vietnam} \\%
\and 
Khanh-Duy Le\thanks{e-mail: lkduy@fit.hcmus.edu.vn}\\%
    \parbox{1.4in}{\scriptsize \centering University of Science \\ VNU-HCM, Vietnam} \\%
\and
Tam V. Nguyen\thanks{e-mail: tamnguyen@udayton.edu}\\%
    \parbox{1.4in}{\scriptsize \centering Dept. of Computer Science \\ University of Dayton, US} \\%
\and 
Minh-Triet Tran\thanks{e-mail: tmtriet@fit.hcmus.edu.vn}\\%
    \parbox{1.4in}{\scriptsize \centering University of Science \\ VNU-HCM, Vietnam} \\%
\and 
Trung-Nghia Le\thanks{e-mail: ltnghia@fit.hcmus.edu.vn (Corresponding author)}\\%
    \parbox{1.4in}{\scriptsize \centering University of Science \\ VNU-HCM, Vietnam} \\%
}

\abstract{
The fashion e-commerce industry has witnessed significant growth in recent years, prompting exploring image-based virtual try-on techniques to incorporate Augmented Reality (AR) experiences into online shopping platforms. However, existing research has primarily overlooked a crucial aspect - the runtime of the underlying machine-learning model. While existing methods prioritize enhancing output quality, they often disregard the execution time, which restricts their applications on a limited range of devices. To address this gap, we propose Distilled Mobile Real-time Virtual Try-On (DM-VTON), a novel virtual try-on framework designed to achieve simplicity and efficiency. Our approach is based on a knowledge distillation scheme that leverages a strong Teacher network as supervision to guide a Student network without relying on human parsing. Notably, we introduce an efficient Mobile Generative Module within the Student network, significantly reducing the runtime while ensuring high-quality output. Additionally, we propose Virtual Try-on-guided Pose for Data Synthesis to address the limited pose variation observed in training images. Experimental results show that the proposed method can achieve 40 frames per second on a single Nvidia Tesla T4 GPU and only take up 37 MB of memory while producing almost the same output quality as other state-of-the-art methods. 
DM-VTON stands poised to facilitate the advancement of real-time AR applications, in addition to the generation of lifelike attired human figures tailored for diverse specialized training tasks.
} 

\CCScatlist{
  \CCScatTwelve{Mixed/augmented reality}{Real-time systems}{Virtual try-on}{Knowledge distillation}
}

\begin{document}



\maketitle

\section{Introduction}


In recent years, the fashion industry, particularly fashion e-commerce, has witnessed significant advancements. Despite these improvements, customers still face the limitation of having to physically visit stores to try on their wanted clothes. As a result, there is a growing interest in virtual try-on (VTON)~\cite{Ge-CVPR2021-Parser, Song-2022ISMAR-VTONShoes, He-CVPR2022-Style}, which demonstrates the potential to enhance shopping experiences by integrating Augmented Reality (AR). Furthermore, VTON exhibits the latent capability to engender new data from practical contexts to support diverse AI applications. We can also utilize them to bring new views to be overlaid on reality for immersive experience.


\begin{figure}[t!]
  \centering
  \includegraphics[width=\columnwidth]{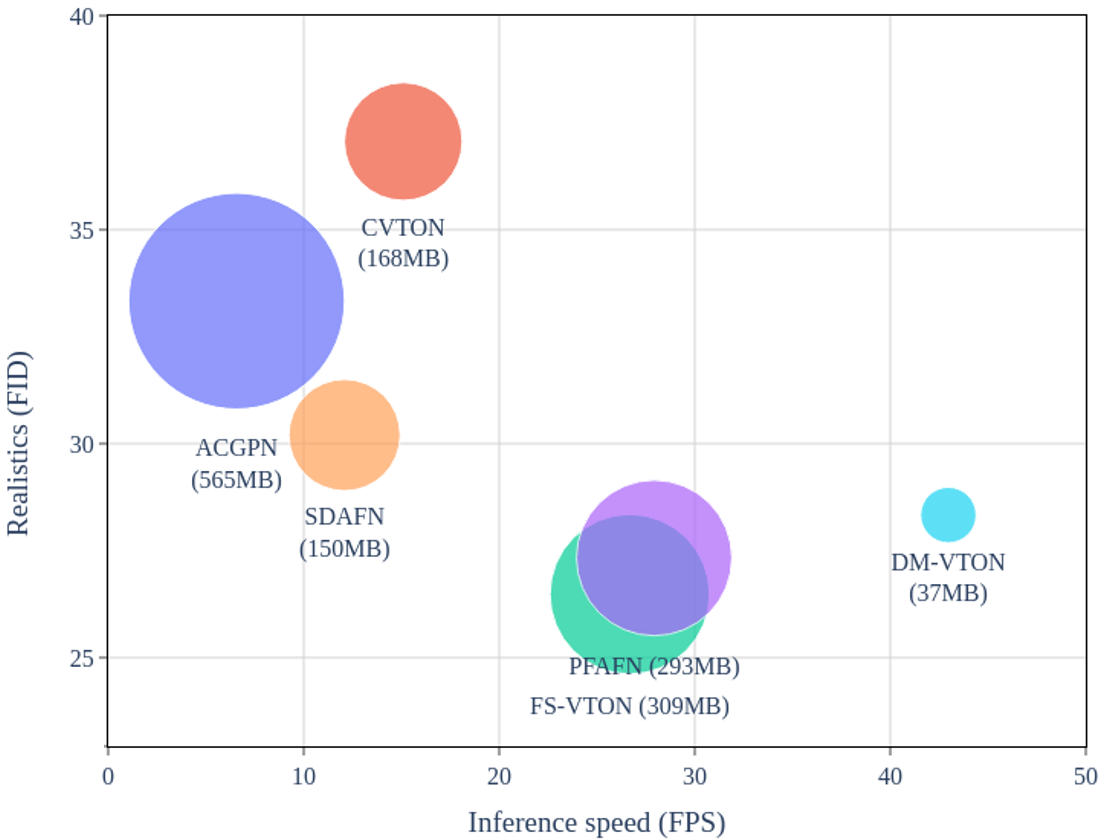}
  \caption{The comparison of our method (DM-VTON) and SOTA methods on VITON test set~\cite{Han-CVPR2018-Viton} in terms of realistic results (FID~\cite{Heusel-NeurIPS2017-FID}, lower is better), inference speed (FPS, higher is better), and memory usage. The size of each bubble represents the memory footprint. FPS is measured using a single Nvidia Tesla T4 GPU.}
  \label{fig:teaser}
  \vspace{-5mm}
\end{figure}

To the best of our knowledge, existing works on image-based VTON do not put their concern about the complexity of their models. They depend on many information like the human semantic segmentation map (body-parser map) and the human pose, denoted as \emph{human representation}, to enhance the output quality. As a result, they take too much time for inference, preventing them from being applied in real-time applications (ACGPN~\cite{Wang-ECCV2018-Toward}, SDAFN~\cite{Bai-ECCV2022-Single} and C-VTON~\cite{Fele-CVPRW2022-CVTON} as in~\autoref{fig:teaser}). Recent methods~\cite{Issenhuth-ECCV2020-Do, Ge-CVPR2021-Parser} have removed the dependency on human representation, thus, reducing the run time considerably. However, they still suffer from using large memory footprints, impacting the requirements for AR devices to run them (PF-AFN~\cite{Ge-CVPR2021-Parser} and FS-VTON~\cite{He-CVPR2022-Style} as in~\autoref{fig:teaser}).

\begin{figure*}[t!]
  \centering
  \includegraphics[width=\textwidth]{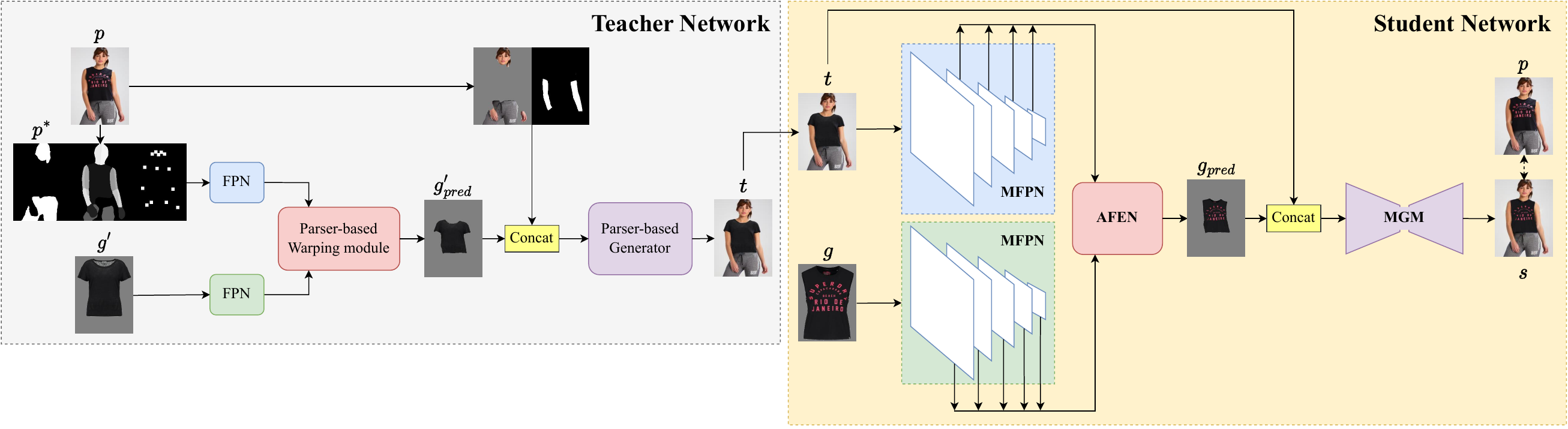}
  \caption{Overview architecture of the proposed Distilled Mobile Real-time Virtual Try-On (DM-VTON) framework. The parser-based Teacher network generates a synthetic image as the input for training the Student network.
  }
  \label{fig:mobile-tryon}
  \vspace{-2mm}
\end{figure*}

Addressing those issues, we propose a novel framework, \textbf{D}istilled \textbf{M}obile real-time \textbf{V}irtual \textbf{T}ry-\textbf{ON} \textbf{(DM-VTON)}, to achieve faster run time and less memory consumption while also producing results of the same quality. The framework consists of two networks: a Teacher network and a Student network. The Teacher network serves as the source of information and guides the Student network through Knowledge Distillation~\cite{Hinton-Arxiv2015-Distilling}. Specifically, the Teacher is trained with the objective of generating the try-on result from the person image, target garment, and human representation (parser-based approach). Because we only use the Teacher during the training phase, we build it using a SOTA VTON architecture~\cite{He-CVPR2022-Style}. Taking advantage of the synthetic result and the original garment, the Student can learn to reconstruct the original realistic image without the need for human representation (parser-free approach). As we use the Student network for inference, we consider the trade-off between its runtime and output quality. We introduce two components in the Student network: the Mobile Feature Pyramid Network (MFPN) and Mobile Generative Module (MGM). Both components are based on the lightweight MobileNet~\cite{Sandler-CVPR2018-Mobilenetv2} architecture, which has been proven to achieve higher throughput and smaller memory usage while producing comparable results to other architectures in the same work. Therefore, our proposed DM-VTON exhibits promising prospects in the realm of real-time AR applications, alongside the generation of verisimilar human attire for the purpose of specialized task-oriented training paradigms.

Besides, common VTON datasets~\cite{Han-CVPR2018-Viton, Morelli-CVPR2022-Dress} only contain a limited range of pose variations. That makes the models trained on those datasets suffer from overfitting. As a result, they perform poorly in real-life scenes. To overcome this problem, we introduce \textbf{V}irtual \textbf{T}ry-on-guided \textbf{P}ose for \textbf{D}ata \textbf{S}ynthesis (VTP-DS), an automatic pipeline to enrich the diversity of the poses in the mentioned datasets. The pipeline has two key ideas: self-checking the results by calculating the Object Keypoint Similarity (OKS)~\cite{Lin-ECCV2014-Microsoft}, and synthesizing new images by using a diffusion network~\cite{Bhunia-CVPR2023-Person}. Given a VTON framework, the pipeline can automatically identify input images where the framework generates results with incorrect poses and extract the corresponding pose information. Then those poses are used as guidance to synthesize new person images from a single image of that person. By utilizing the VTP-DS pipeline, we can enrich the datasets with a wider range of pose variations, thus enhancing the training process and improving the performance of VTON models in real-life scenarios.

We conducted experiments comparing our DM-VTON framework with other SOTA methods in terms of inference speed, memory usage, and the realisticness of the output. During the experimentation, we carefully evaluated the trade-off between those factors. As depicted in~\autoref{fig:teaser}, our DM-VTON framework outperforms all existing SOTA methods regarding inference speed and memory usage while maintaining an equal quality of results. In summary, our contributions are as follows:
\begin{itemize}
    \vspace{-2mm}
    \item To address the limitation of image-based VTON that concerns inference time and memory consumption, we propose the DM-VTON framework based on knowledge distillation learning. By developing a lightweight Student network, we can reduce the number of floating point operations (FLOPs) and parameters of the model, thus making it easier to deploy and operate on AR devices.
    \vspace{-2mm}
    \item We introduce VTP-DS, an automatic fashion-pose data generation pipeline designed to enrich existing fashion datasets by synthesizing new person poses from a single image of that person. The pipeline utilizes a VTON framework to identify challenging poses and subsequently generates additional images for those specific poses. These generated images are then utilized in the training process, improving the framework's performance in real-world scenarios.
    \vspace{-2mm}
    \item Experiment results show that DM-VTON achieves faster inference and more efficient resource utilization while producing comparable result quality with existing SOTA methods, which proves the efficiency of our proposed framework.
\end{itemize}

\section{Related work}

Image-based VTON techniques can be classified into two categories: parser-based and parser-free approaches. Both of them typically involve three steps: extracting the intrinsic input features, warping the input garment to fit the clothing area of the person image, and performing the replacement using a generative model.

As for parser-based VTON methods, they require human representation, including the body-parser map and human pose, to calculate the warping transformation matrix. The very first methods that pave for this approach are VITON~\cite{Han-CVPR2018-Viton} and CP-VTON~\cite{Wang-ECCV2018-Toward}. To improve the output quality, ACGPN~\cite{Yang-CVPR2020-Towards} also warps the body-parser map along the target garment. SDAFN~\cite{Bai-ECCV2022-Single} reduces the need for the parser map but still needs the human pose, though. Recently, ClothFlow~\cite{Han-ICCV2019-Clothflow} is the first method that uses the appearance flow to guide the warping procedure, and this approach is still used in current SOTA methods~\cite{Ge-CVPR2021-Parser, He-CVPR2022-Style}.

On the other hand, the parser-free approach only requires an input garment and a person image for inference. Thus, this makes the inference process much faster and more independent from other intermediate models. WUTON~\cite{Issenhuth-ECCV2020-Do}, the pioneering parser-free method, produces noticeable artifacts due to using the same input-output pairs for both Teacher and Student networks~\cite{Ge-CVPR2021-Parser}. Addressing that issue, PF-AFN~\cite{Ge-CVPR2021-Parser} introduces a new Knowledge Distillation-based training pipeline, in which the Student network takes the Teacher output as its input and has its output supervised by the original images. This training methodology has become the standard for subsequent parser-free methods. RMGN~\cite{Lin-IJCAI2022-RMGN} improves the generation part by using SPADE blocks~\cite{Park-CVPR2019-Semantic}, while FS-VTON~\cite{He-CVPR2022-Style} improves the warping part by using StyleGan blocks~\cite{Karras-CVPR2019-Style}. 

There are also methods that aim to perform VTON on a sequence of frames. These methods use techniques like memory-based~\cite{Zhong-ACMMM2021-Mvton} or optical flow~\cite{Kuppa-WACV2021-ShineOn} to keep the temporary consistency between the frames. However, these methods are still based on the parser-based approach, which takes considerable time to calculate the human representation.

In this paper, we adopt the parser-free approach to prioritize speed, as calculating human representation is a time bottleneck in the try-on process. However, we take a distinct step from existing methods by modifying the Student network for improved speed and reduced memory consumption while keeping the parser-based approach in the Teacher network to preserve the output quality.

\begin{figure}[t!]
  \centering
  \includegraphics[width=0.93\columnwidth]{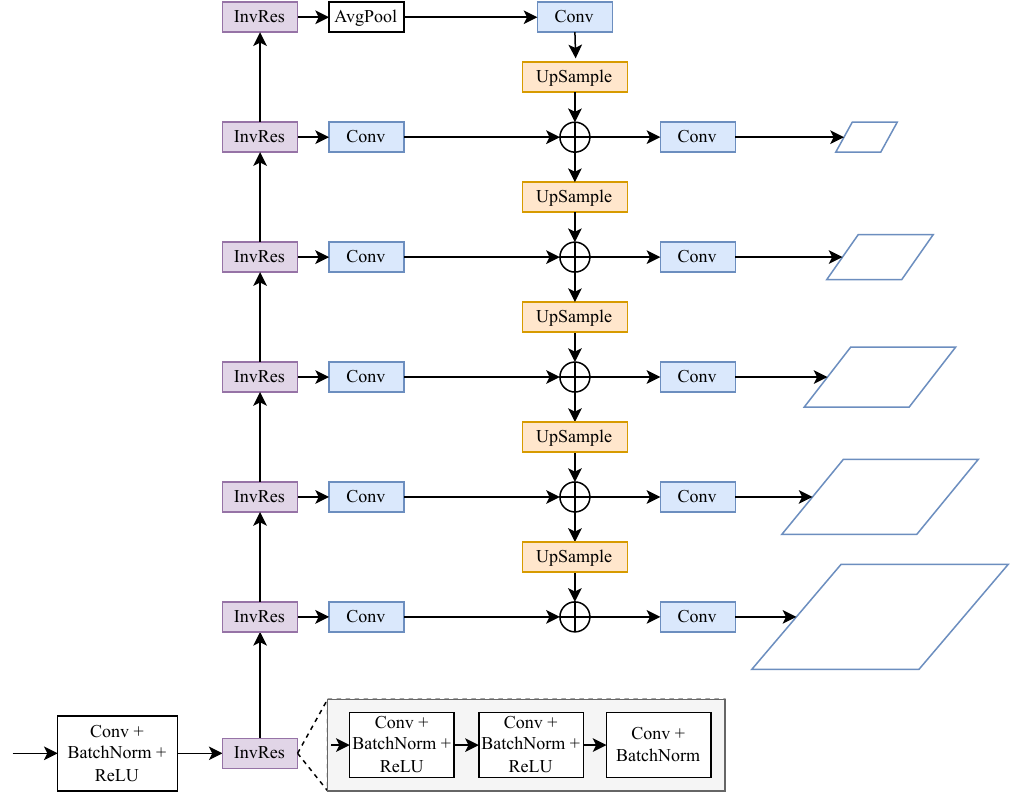}
  \caption{Mobile Feature Pyramid Network architecture}
  \label{fig:mobile-pyramid}
  \vspace{-5mm}
\end{figure}

\section{Method}

\subsection{Overview}

Our objective is to generate an image of a person wearing a specific garment while preserving the rest. To achieve this goal, we adopt the knowledge distillation training pipeline~\cite{Hinton-Arxiv2015-Distilling, Issenhuth-ECCV2020-Do, Ge-CVPR2021-Parser, He-CVPR2022-Style, Lin-IJCAI2022-RMGN} to develop a Distilled Mobile Real-time Virtual Try-On (DM-VTON) framework (see~\autoref{fig:mobile-tryon}). Our proposed DM-VTON consists of two networks: Teacher and Student networks. Both include three main components: feature extractor, clothes-warping module, and generator. 

The Teacher network aims to produce the VTON result using the parser-based training process. The Student network then utilizes the Teacher network to generate synthetic input images, enabling the Student network to be supervised by the original images without relying on human representation. To ensure high-quality output, the Teacher network is built upon SOTA VTON models. As we focus on inference speed, we propose lightweight components for the Student network. 

\subsection{Teacher Network}

The main purpose of this network is to generate a synthetic person image that serves as the input for the Student training process. Furthermore, the Teacher also helps this process through a knowledge distillation scheme. In particular, we take advantage of the SOTA method of VTON: FS-VTON \cite{He-CVPR2022-Style}. As shown in~\autoref{fig:mobile-tryon}, it incorporates two feature pyramid networks (FPN)~\cite{Lin-CVPR2017-FPN}, enabling the extraction of features from the human representation $p^*$ and garment image $g'$. To achieve the garment deformation functionality, the Teacher network utilizes a style-based global appearance flow estimation network that uses modulated convolution \cite{Karras-CVPR2019-Style}. This network first predicts a coarse appearance flow via extracted global style vector and then refines it locally to capture the global and local correspondence between the garment and the target person. This makes the Teacher network more robust against the problems of detail-preserving and large misalignment. Finally, the warped clothes and the preserved region on the human body are concatenated as the input for try-on result generation. The generator of our Teacher network follows the encoder-decoder architecture with skip connections, which have been proven effective in detail preservation.

Because the inputs of the parser-based model (i.e. human representation) contain more semantic information when compared to those in the parser-free model, we employ an adjustable knowledge distillation learning scheme~\cite{Ge-CVPR2021-Parser} with a distillation loss to guide the feature extractor of the Student network:
\begin{equation} 
    L_{dis}=\psi \sum_{i=1}^N\|t_{p_i}-s_{p_i}\|_2,
\end{equation}

\begin{equation} 
    \psi= \{\begin{array}{l} 1, ~if~ \|t-p\|_1<\|s-p\|_1 \\ 0,~otherwise \end{array},
\end{equation}
where $t_{p_i}$ and $s_{p_i}$ denote the feature maps at the $i$-th scale extracted from $p^*$ within the Teacher network and the synthetic image $t$ within the Student network; $t$, $s$ are the try-on result of the Teacher and Student, respectively; $p$ is the person image ground truth.
\subsection{Student Network}

We propose a parser-based approach for synthesizing try-on images with increased speed compared to previous methods while ensuring accuracy. As shown in~\autoref{fig:mobile-tryon}, our Student network consists of three key components: Mobile Feature Pyramid Network (MFPN), Appearance Flow Estimation Network (AFEN), and Mobile Generative Module (MGM). These components synergistically collaborate to extract features, manipulate garments through deformation, and generate try-on images. The AFEN introduced by Ge et al.~\cite{Ge-CVPR2021-Parser} proved effective in deforming garments by using appearance flow. The MFPN and MGM are built upon the architecture of MobileNetV2~\cite{Sandler-CVPR2018-Mobilenetv2} with Inverted Residual blocks specifically designed to optimize computational efficiency and model size.

\begin{figure}[t!]
  \centering
  \includegraphics[width=0.93\columnwidth]{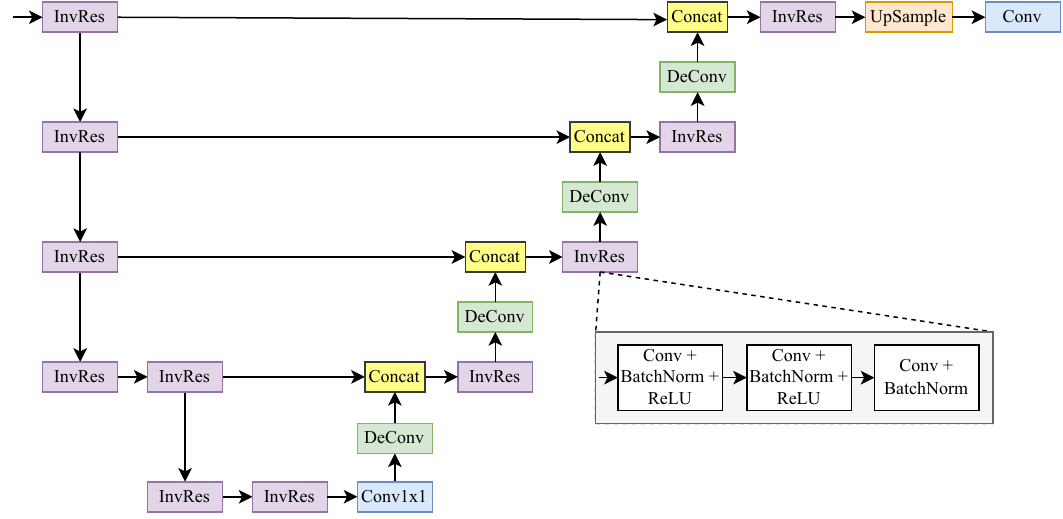}
  \caption{Mobile Generative Module architecture}
  \label{fig:mobile-unet}
  \vspace{-5mm}
\end{figure}

\subsubsection{Mobile Feature Pyramid Network}
As shown in~\autoref{fig:mobile-pyramid}, MFPN incorporates the architecture of MobileNetV2 with Inverted Residual blocks~\cite{Sandler-CVPR2018-Mobilenetv2} to a Feature Pyramid Network. By leveraging the capabilities of two MFPN blocks, we extract two-branch N-level feature maps from person and garment images within a parser-free network. These features are fed into the Appearance Flow Estimation Network (AFEN) to predict the appearance flow map for garment deformation.

\subsubsection{Appearance Flow Estimation Network}
This component aims to deform the garment to fit the human pose while preserving the texture. Following the work of Ge et al.~\cite{Ge-CVPR2021-Parser}, we adopt an appearance flow estimation network (AFEN) comprising subnetworks equipped with varying sizes of convolution layers. These subnetworks are responsible for estimating flows based on extracted multi-level feature maps. The outcome of this network can capture the long-range correspondence between the garment image and the person image, effectively minimizing issues related to misalignment. To enhance the preservation of clothing characteristics, this module is optimized with the second-order smooth loss:
\begin{equation} 
L_{sec}=\sum_{i=1}^N \sum_t \sum_{\pi \in N_t} CharLoss\left(f_i^{t-\pi}+f_i^{t+\pi}-2 f_i^t\right),
\end{equation}
where $f_i^t$ denotes the $t$-th point on the $i$-th scale flow map; $N_t$ is the set of horizontal, vertical, and diagonal neighborhoods around the $t$-th point; and $CharLoss$ denotes generalized Charbonnier loss~\cite{Sun-IJCV2014-Quantitative}.

\begin{figure*}[t!]
 \centering 
 \includegraphics[width=0.7\textwidth]{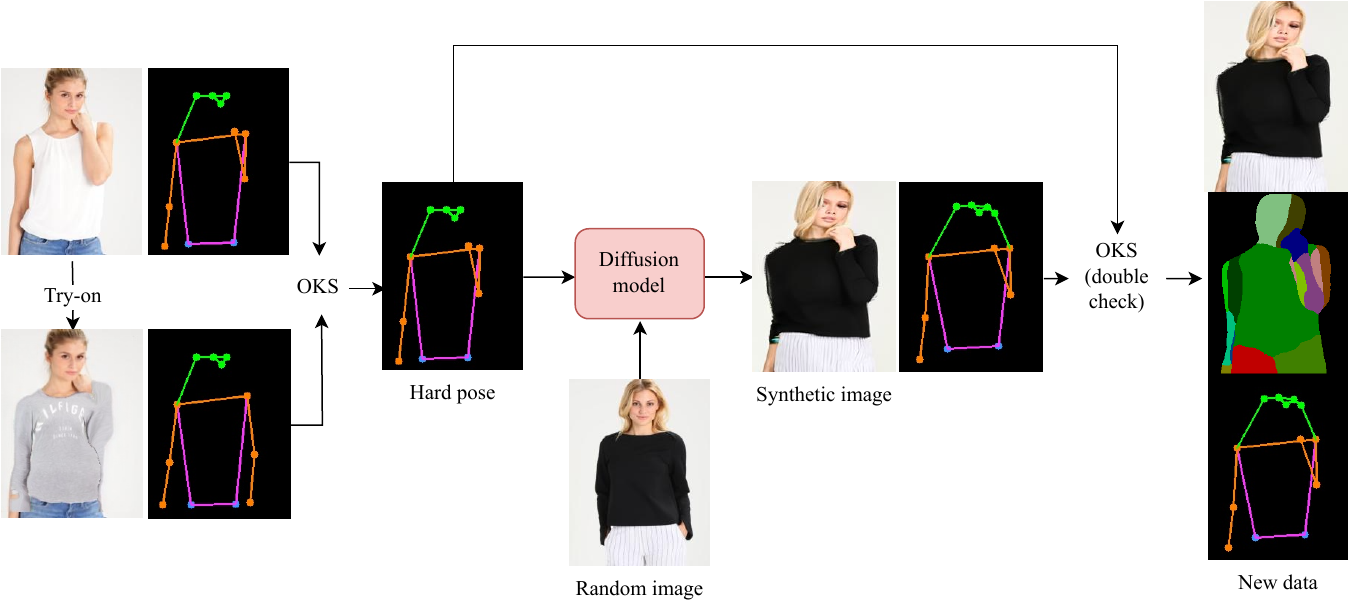}
  \caption{Overview of Virtual Try-on-guided Pose for Data Synthesis pipeline.}
   \label{fig:augment-pipeline}
   \vspace{-2mm}
\end{figure*}

\subsubsection{Mobile Generative Module}
To synthesize the entire try-on image from the warped image and target person image, we develop a Mobile Generative Module, the integration of the architectural principles of UNet~\cite{Ronneberger-MICCAI2015-Unet} and MobileNetV2~\cite{Sandler-CVPR2018-Mobilenetv2} as illustrated in~\autoref{fig:mobile-unet}. The primary objective behind the design of this generator is to reduce both the computational burden and the model's overall size.


\subsubsection{Loss Function}
During training, we optimize the warping module separately in the first stage and then train together with the generator in the last stage. The loss function used in the first stage is defined as:
\begin{equation}
    L^{warp} = \lambda^{warp}_{l}L^{warp}_{l} + \lambda^{warp}_{per}L^{warp}_{per} + \lambda_{sec}L_{sec} + \lambda_{dis}L_{dis},
\end{equation}
where $ L^{warp}_{l} = \|g_{pred} - p \odot m_{gt}\|$ denotes pixel-wise L1 loss; $L^{warp}_{per} = \sum_i{\|\Phi_i(g_{pred}) - \Phi_i(p \odot m_{gt})\|}$ is the perceptual loss~\cite{Johnson-ECCV2016-Perceptual}; $L^{warp}_{sec}$ is the second-order smooth loss; $L^{warp}_{dis}$ is the distillation loss; $g_{pred}$ is warped garment.  $p$ is the person image ground truth with the garment mask $m_{gt}$; $\Phi_i$ is the $i$-th block of pre-trained VGG19~\cite{Simonyan-ArXiv2014-VGG}.

With the generative module, we also apply L1 loss perceptual loss~\cite{Johnson-ECCV2016-Perceptual} between the synthesized image and the ground truth image to supervise the training process of MGM:
\begin{equation}
    L^{gen} = \lambda^{gen}_{l}L^{gen}_{l} + \lambda^{gen}_{per}L^{gen}_{per},
\end{equation}
where $ L^{gen}_{l} = \|s - p\|$ is L1 loss and $L^{gen}_{per} = \sum_i{\|\Phi_i(s) - \Phi_i(p)\|}$ is the perceptual loss~\cite{Johnson-ECCV2016-Perceptual}. $s$ and $p$ are the generated output of the Student network and the person image ground truth, respectively. 


\subsection{Virtual Try-on-guided Pose for Data Synthesis}
\begin{figure}[t!]
    \vspace{-3mm}
    \centering
    \subfloat[VITON]{\includegraphics[width=0.47\columnwidth]{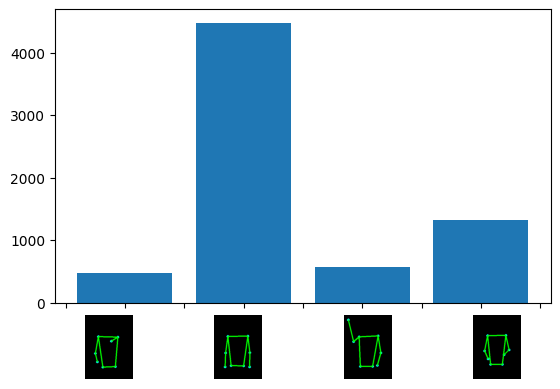}}
    \subfloat[VITON + synthesized data]{\includegraphics[width=0.47\columnwidth]{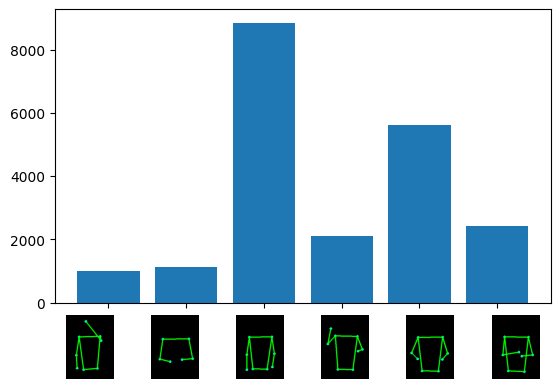}} 
    \caption{Pose distribution in VITON dataset~\cite{Han-CVPR2018-Viton}.}
    \label{fig:pose-distribution}
    \vspace{-5mm}
\end{figure}

By using the K-Means clustering algorithm, we observe that the original VITON dataset~\cite{Han-CVPR2018-Viton} is mainly composed of images with straight-arm poses (as in~\autoref{fig:pose-distribution}(a)). This bias creates a challenge as models trained on such data are prone to overfit and perform poorly on images with different upper-body poses. To tackle this problem, we propose the Virtual Try-on-guided Pose for Data Synthesis (VTP-DS) pipeline. Intending to improve the existing VTON framework, the pipeline incorporates two key ideas: automatically detecting poorly performed poses using the Object Keypoint Similarity (OKS)~\cite{Lin-ECCV2014-Microsoft} and synthesizing new training data targeting those poses. The overview of the VTP-DS pipeline is illustrated in~\autoref{fig:augment-pipeline}.

Given a person image, we extract that person's pose using the YOLOv7 pose estimation method~\cite{Wang-CVPR2023-YOLOv7}. Then, we utilize our trained DM-VTON model to perform VTON on the input image. The extracted pose from the resulting image is compared with the pose of the input image using a customized OKS~\autoref{formulat:oks}.  
\begin{equation}
\centering
\frac{1}{|P|}\sum_{i \in P} \exp(\frac{-d_i^2}{2 s^2 k_i^2}),
\label{formulat:oks}
\end{equation}
where $P$ denotes the set of arm and hand keypoints; $d_i$ denotes the Euclidean distance between the keypoint $i$ of two poses; $s$ denotes the total area containing the pose; $k_i$ is the constant provided by Lin et al.~\cite{Lin-ECCV2014-Microsoft} to represent the standard deviation for keypoint $i$. Only the arm and hand keypoints contribute to the formula as we focus on distinguishing different upper-body poses. If the OKS score falls below a specified threshold $t=0.9$, it is identified as a hard pose.

\begin{table*}[t!]
    \centering
    \caption{Quantitative results between DM-VTON and SOTA VTON methods. The $\dagger$ marker indicates the results measured by the generated images provided by the authors. The speed was evaluated on a single Nvidia T4 GPU.}
    \vspace{-3mm}
    \label{table:tryon-speed}
    \scriptsize
    \begin{tabular}{llcccccccc}
        \toprule
        \textbf{Method}  & \textbf{Published} & Parser & Pose & \textbf{FID $\downarrow$} & \textbf{LPIPS $\downarrow$} & \textbf{Runtime (ms) $\downarrow$} & \textbf{FLOPs (B) $\downarrow$} & \textbf{Memory usage (MB) $\downarrow$}\\
        \midrule
        ACGPN~\cite{Yang-CVPR2020-Towards} & CVPR 2020 & \checkmark & \checkmark & 33.33 & 0.231 & 153.64 & 399.08 & 565.86 \\
        PF-AFN~\cite{Ge-CVPR2021-Parser} & CVPR 2021 & & & 27.33 & 0.216 & 35.80 & 137.85 & 293.25 \\
        C-VTON$\dagger$~\cite{Fele-CVPRW2022-CVTON} & CVPRW 2022 & \checkmark & & 37.06 & 0.241 & 66.90 & 108.47 & 168.60 \\
        SDAFN~\cite{Bai-ECCV2022-Single} & ECCV 2022 & & \checkmark & 30.20 & 0.245 & 83.42 & 149.40 & 150.87 \\
        FS-VTON~\cite{He-CVPR2022-Style} & CVPR 2022 & & & 26.48 & 0.200 & 37.49 & 132.98 & 309.25 \\
        \midrule
        \textbf{DM-VTON} & Ours & & & 28.33 & 0.215 & 23.27 & 69.82 & 37.79 \\
    \bottomrule
\end{tabular}
\end{table*}

\begin{figure*}[t!]
  \centering
  \includegraphics[width=\textwidth]{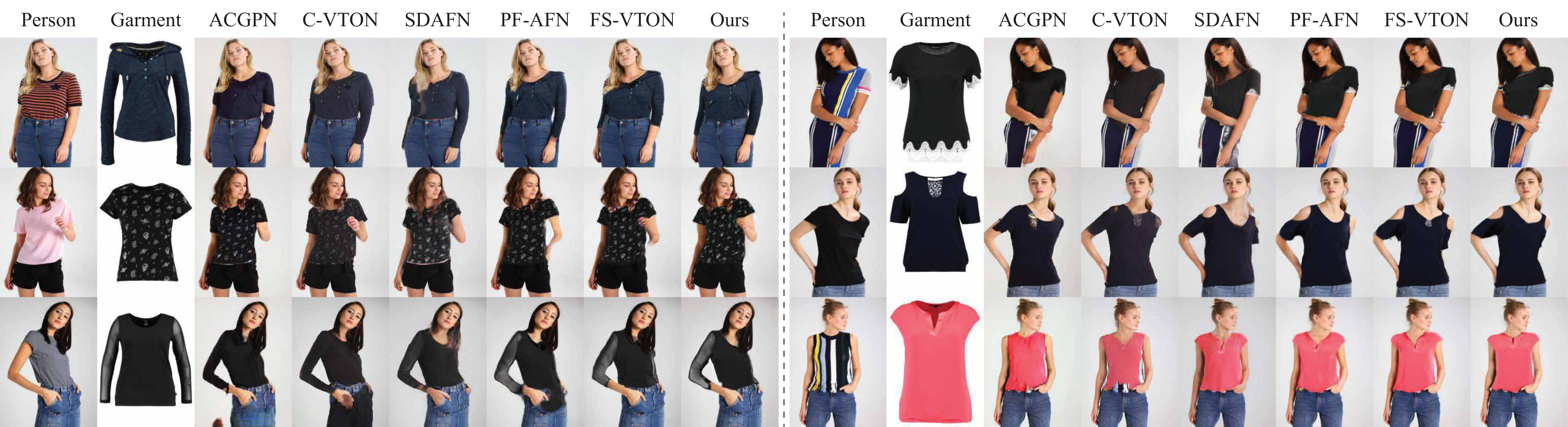}
  \caption{Qualitative comparison on VITON-Clean dataset~\cite{Han-CVPR2018-Viton}.}
  \label{fig:qualitative-viton}
  \vspace{-2mm}
\end{figure*}

Once identifying a hard pose, we randomly pick a person image from the VITON dataset. Then it leverages Bhunia's Diffusion model~\cite{Bhunia-CVPR2023-Person} to synthesize a new image of the person in the corresponding pose. To ensure the accuracy of the synthesized image, we perform a double-check using the OKS to verify the correctness of the output pose. Finally, DensePose~\cite{Guler-CVPR2018-Dense} is utilized to generate the body-parser map of the synthesized image.

We initially collected videos from Youtube to synthesize additional data for training networks, specifically focusing on posing or catwalk videos. These videos had varying durations, ranging from 1 to 10 minutes. Subsequently, we extracted individual frames from these videos, which served as the input for our VTP-DS pipeline. 
After that, we manually removed low-quality results, resulting in 14,314 high-quality synthesized images for training the networks. 

To access the quality of synthesized images, we use the K-means algorithm combined with our modified OKS. As details of the pose clustering results shown in~\autoref{fig:pose-distribution}, data in the VITON training set mainly focuses on poses with simple poses (i.e. arms are less covered, low rotation amplitude). Meanwhile, when combined with our synthesized images, new pose clusters appear, and the concentration of data in groups is less imbalanced, which can help to train robust VTON models. 

\section{Experiments}


\subsection{Experimental Settings}

VITON~\cite{Han-CVPR2018-Viton}, the most popular dataset for evaluating VTON, was used to evaluate methods. It contains 16,253 frontal-view upper-body woman and top clothing image pairs with $256 \times 192$ resolution. However, we followed the work of Han et al.~\cite{Han-ICCV2019-Clothflow} to filter out duplicates and ensure no data leakage happens, remaining 6,824 training image pairs and 416 testing image pairs in the cleaned VITON dataset, denoted by VTION-Clean. We combine the VTION-Clean training set and our synthesized images to train our proposed method.


Fréchet Inception Distance (FID)~\cite{Heusel-NeurIPS2017-FID} and Learned Perceptual Image Patch Similarities (LPIPS)~\cite{Zhang-CVPR2018-LPIPS} metrics were used to evaluate the similarity of try-on results to real images. 


\subsection{Experimental Results}
We compared the performance of our proposed MD-VTON with several SOTA methods, including ACGPN~\cite{Yang-CVPR2020-Towards}, PF-AFN~\cite{Ge-CVPR2021-Parser}, C-VTON~\cite{Fele-CVPRW2022-CVTON}, SDAFN~\cite{Bai-ECCV2022-Single}, FS-VTON~\cite{He-CVPR2022-Style}. Comparison of MD-VTON against those methods in terms of image quality (i.e. FID, LPIPS), inference speed (ms), FLOPs (Floating point operations), and memory usage (MB) is shown in~\autoref{table:tryon-speed}. Our proposed method outperforms other SOTAs in terms of runtime, FLOPs, and memory consumption. On the other hand, our DM-VTON achieves slightly higher FID and LPIPS scores than those of PF-AFN~\cite{Ge-CVPR2021-Parser} and FS-VTON~\cite{He-CVPR2022-Style}. The experimental results prove that the proposed DM-VTON can run in real-time (i.e. 43 frames per second) with small memory consumption but still retains high-quality VTON results. The visualization of compared methods is illustrated in~\autoref{fig:qualitative-viton}.

\section{Pilot Study}
\begin{figure}[t!]
    \centering
    \subfloat[Human model samples]{\includegraphics[width=0.9\columnwidth]{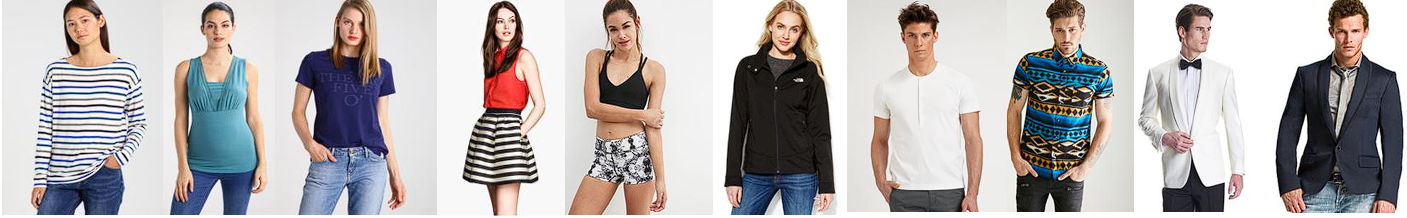}} \\
    \vspace{-4mm}
    \subfloat[Garment samples]{\includegraphics[width=0.9\columnwidth]{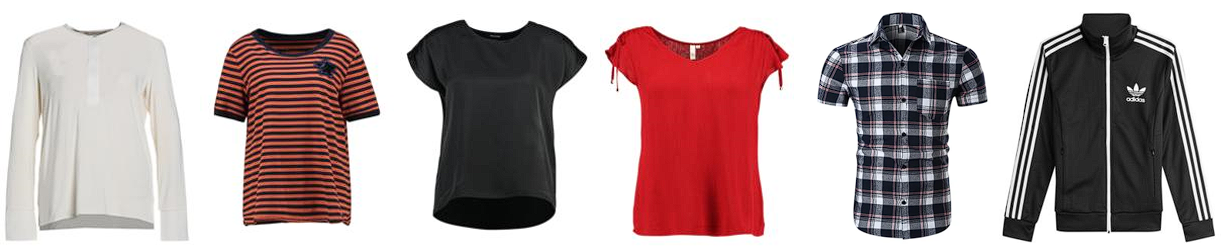}}
    \caption{Examples of data used in our pilot study.}
    \label{fig:pilot-study-data}
    \vspace{-2mm}
\end{figure}

We invited 12 participants who are university students and researchers in the 18-44 age range. We let the users experience our DM-VTON and collected their feedback. The experimental scenario was that while shopping online at home, the users come across a particular garment that catches their eyes, but they are unsure whether it looks good on them. We offered them an application to try on such garments before making the purchasing decision. Specifically, we prepared a collection of 20 garments taken from the VITON~\cite{Han-CVPR2018-Viton} and PolyvoreOutfits~\cite{Han-ACMMM2017-Learning} datasets (see~\autoref{fig:pilot-study-data}). These garments were carefully selected to represent a variety of colors, shapes, and textures.

Each participant took part in a 10-minute session in which the participant was asked to perform a VTON using our provided model images and VTON on themselves directly captured from our camera. In terms of human models, we used 10 person images in the VITON~\cite{Han-CVPR2018-Viton} and DeepFashion~\cite{Liu-CVPR2016-DeepFashion} datasets (see~\autoref{fig:pilot-study-data}).

Upon completing the trial, we interviewed participants and asked for their feedback. Our primary objective was to evaluate how our system influenced their purchasing decisions. We also gathered their feedback about the output quality and whether they prefer using the model images or their own images to enhance the overall user experience in the future.

\begin{figure}[t!]
  \centering
  \includegraphics[width=0.72\columnwidth]{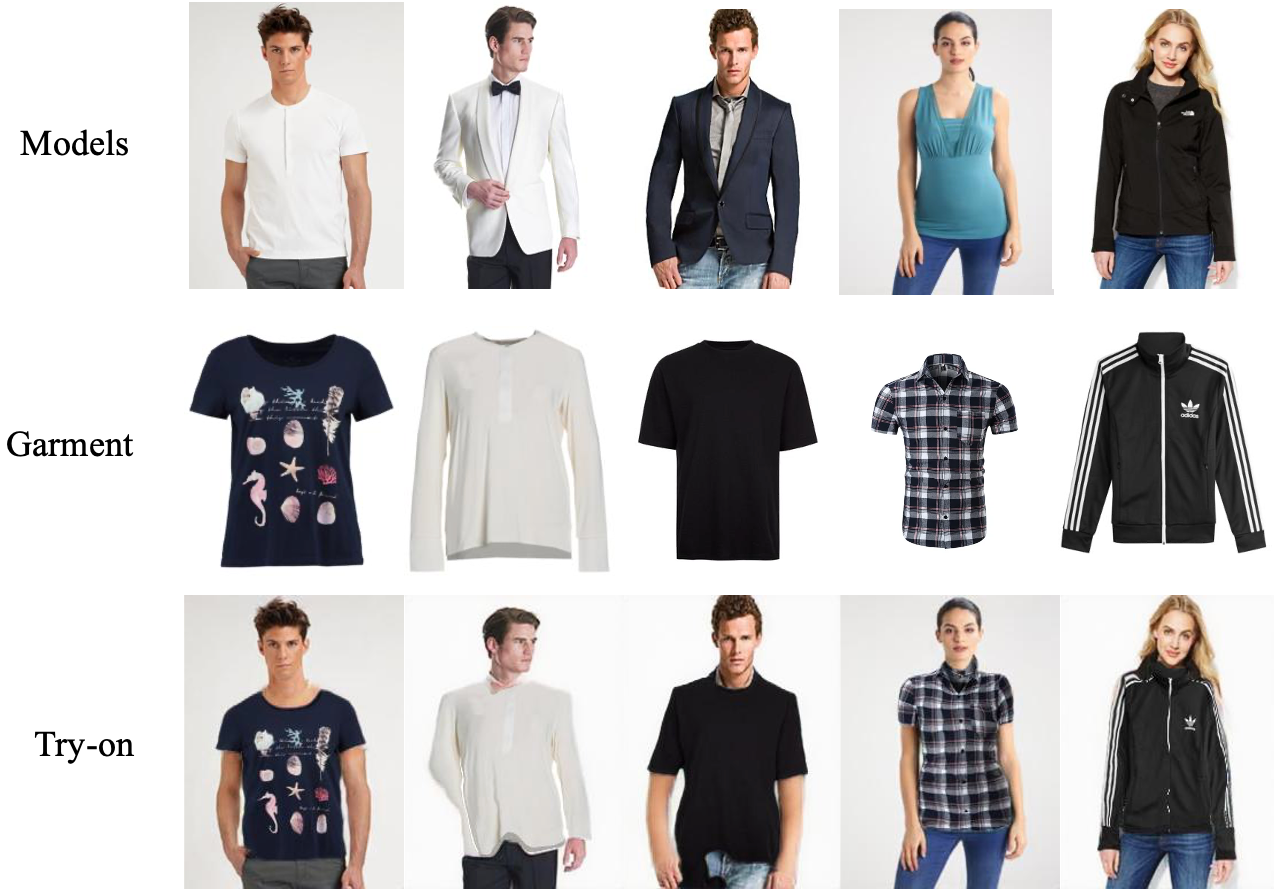}
  \caption{Results obtained when users performed VTON on our provided human models in the pilot study.}
  \label{fig:study-result}
  \vspace{-5mm}
\end{figure}

Most of the participants agreed that trying on clothes in various poses helped them visualize the suitability of the garments before making a purchase decision. Specifically, $66.7\%$ of the participants felt confident enough to make the purchasing decision after using our system, while the remaining $33.3\%$ had doubts about the truthiness of the models. Moreover, $83.3\%$ of the participants preferred using their images for try-on, as it provided a more realistic experience for them. On the other hand, the remaining $16.7\%$ considered both options, as the provided models allowed them to see the best representation of the garments, such as with appropriate brightness and poses.~\autoref{fig:study-result} illustrates some VTON results on our provided models. Due to privacy issues, we did not capture the VTON results on the participants' images.

We also received valuable feedback from participants on areas for improvement. In real-life conditions, the background, brightness, and quality of the captured images might not be suitable for trying on clothes, which is due to the fact that our train and test datasets only contain simple backgrounds and have proper brightness conditions. Thus, applying pre-processing techniques, such as segmentation and brightness equalization, is necessary to address this issue. Additionally, participants also suggested that our system exhibited inconsistencies when dealing with complex poses such as half-turn or crossed-arm poses. Their feedback is useful in enhancing the user experience in the future.

\section{Conclusion}
We present the simple yet efficient DM-VTON framework. By leveraging the knowledge distillation scheme, we developed a lightweight parser-free network to boost the processing speed. Our proposed network utilizes mobile-based architectures, resulting in achieving real-time VTON capabilities while maintaining high-quality output and computational efficiency. In addition, the Teacher network, trained using a parser-based approach, provides supervision to the Student network, enabling it to learn without relying on the human representation of ground truth. Additionally, to address the limited pose variation observed in the training images, we introduced the VTP-DS to enrich the diversity of poses in the training data. It automatically identifies input images with incorrect poses generated by the framework and generates additional images for those specific poses. 

Experimental results showcase the potential of our method. It can be applied to real-time AR applications, paving the way for improved user experiences in a virtual fashion context. It also can be used to synthesize more data for people with different outfits for training with different specific tasks.

\acknowledgments{
This research was funded by Vingroup and supported by Vingroup Innovation Foundation (VINIF) under project code VINIF.2019.DA19.
}

\bibliographystyle{abbrv-doi}

\bibliography{bibtex}

\begin{thebibliography}{10}

\bibitem{Bai-ECCV2022-Single}
S.~Bai, H.~Zhou, Z.~Li, C.~Zhou, and H.~Yang.
\newblock Single stage virtual try-on via deformable attention flows.
\newblock In {\em ECCV}, pp. 409--425, 2022.

\bibitem{Bhunia-CVPR2023-Person}
A.~K. Bhunia, S.~Khan, H.~Cholakkal, R.~M. Anwer, J.~Laaksonen, M.~Shah, and
  F.~S. Khan.
\newblock Person image synthesis via denoising diffusion model.
\newblock In {\em CVPR}, pp. 5968--5976, 2023.

\bibitem{Fele-CVPRW2022-CVTON}
B.~Fele, A.~Lampe, P.~Peer, and V.~Struc.
\newblock C-vton: Context-driven image-based virtual try-on network.
\newblock In {\em WACV}, pp. 3144--3153, 2022.

\bibitem{Ge-CVPR2021-Parser}
Y.~Ge, Y.~Song, R.~Zhang, C.~Ge, W.~Liu, and P.~Luo.
\newblock Parser-free virtual try-on via distilling appearance flows.
\newblock In {\em CVPR}, pp. 8485--8493, 2021.

\bibitem{Guler-CVPR2018-Dense}
R.~Guler, N.~Neverova, and I.~DensePose.
\newblock Dense human pose estimation in the wild.
\newblock In {\em CVPR}, pp. 18--23, 2018.

\bibitem{Han-ICCV2019-Clothflow}
X.~Han, X.~Hu, W.~Huang, and M.~R. Scott.
\newblock Clothflow: A flow-based model for clothed person generation.
\newblock In {\em ICCV}, pp. 10471--10480, 2019.

\bibitem{Han-ACMMM2017-Learning}
X.~Han, Z.~Wu, Y.-G. Jiang, and L.~S. Davis.
\newblock Learning fashion compatibility with bidirectional lstms.
\newblock In {\em ACM Multimedia}, pp. 1078--1086, 2017.

\bibitem{Han-CVPR2018-Viton}
X.~Han, Z.~Wu, Z.~Wu, R.~Yu, and L.~S. Davis.
\newblock Viton: An image-based virtual try-on network.
\newblock In {\em CVPR}, pp. 7543--7552, 2018.

\bibitem{He-CVPR2022-Style}
S.~He, Y.-Z. Song, and T.~Xiang.
\newblock Style-based global appearance flow for virtual try-on.
\newblock In {\em CVPR}, pp. 3470--3479, 2022.

\bibitem{Heusel-NeurIPS2017-FID}
M.~Heusel, H.~Ramsauer, T.~Unterthiner, B.~Nessler, and S.~Hochreiter.
\newblock Gans trained by a two time-scale update rule converge to a local nash
  equilibrium.
\newblock {\em NeurIPS}, p. 6629–6640, 2017.

\bibitem{Hinton-Arxiv2015-Distilling}
G.~Hinton, O.~Vinyals, and J.~Dean.
\newblock Distilling the knowledge in a neural network.
\newblock {\em arXiv preprint arXiv:1503.02531}, 2015.

\bibitem{Issenhuth-ECCV2020-Do}
T.~Issenhuth, J.~Mary, and C.~Calauzenes.
\newblock Do not mask what you do not need to mask: a parser-free virtual
  try-on.
\newblock In {\em ECCV}, pp. 619--635, 2020.

\bibitem{Johnson-ECCV2016-Perceptual}
J.~Johnson, A.~Alahi, and L.~Fei-Fei.
\newblock Perceptual losses for real-time style transfer and super-resolution.
\newblock In {\em ECCV}, pp. 694--711, 2016.

\bibitem{Karras-CVPR2019-Style}
T.~Karras, S.~Laine, and T.~Aila.
\newblock A style-based generator architecture for generative adversarial
  networks.
\newblock In {\em CVPR}, pp. 4401--4410, 2019.

\bibitem{Kuppa-WACV2021-ShineOn}
G.~Kuppa, A.~Jong, X.~Liu, Z.~Liu, and T.-S. Moh.
\newblock Shineon: Illuminating design choices for practical video-based
  virtual clothing try-on.
\newblock In {\em WACV}, pp. 191--200, 2021.

\bibitem{Lin-IJCAI2022-RMGN}
C.~Lin, Z.~Li, S.~Zhou, S.~Hu, J.~Zhang, L.~Luo, J.~Zhang, L.~li~Huang, and
  Y.~He.
\newblock Rmgn: A regional mask guided network for parser-free virtual try-on.
\newblock In {\em IJCAI}, pp. 1151--1158, 2022.

\bibitem{Lin-CVPR2017-FPN}
T.-Y. Lin, P.~Doll{\'a}r, R.~Girshick, K.~He, B.~Hariharan, and S.~Belongie.
\newblock Feature pyramid networks for object detection.
\newblock In {\em CVPR}, pp. 2117--2125, 2017.

\bibitem{Lin-ECCV2014-Microsoft}
T.-Y. Lin, M.~Maire, S.~Belongie, J.~Hays, P.~Perona, D.~Ramanan,
  P.~Doll{\'a}r, and C.~L. Zitnick.
\newblock Microsoft coco: Common objects in context.
\newblock In {\em ECCV}, pp. 740--755, 2014.

\bibitem{Liu-CVPR2016-DeepFashion}
Z.~Liu, P.~Luo, S.~Qiu, X.~Wang, and X.~Tang.
\newblock Deepfashion: Powering robust clothes recognition and retrieval with
  rich annotations.
\newblock In {\em CVPR}, pp. 1096--1104, 2016.

\bibitem{Morelli-CVPR2022-Dress}
D.~Morelli, M.~Fincato, M.~Cornia, F.~Landi, F.~Cesari, and R.~Cucchiara.
\newblock Dress code: High-resolution multi-category virtual try-on.
\newblock In {\em CVPR}, pp. 2231--2235, 2022.

\bibitem{Park-CVPR2019-Semantic}
T.~Park, M.-Y. Liu, T.-C. Wang, and J.-Y. Zhu.
\newblock Semantic image synthesis with spatially-adaptive normalization.
\newblock In {\em CVPR}, pp. 2337--2346, 2019.

\bibitem{Ronneberger-MICCAI2015-Unet}
O.~Ronneberger, P.~Fischer, and T.~Brox.
\newblock U-net: Convolutional networks for biomedical image segmentation.
\newblock In {\em MICCAI}, pp. 234--241, 2015.

\bibitem{Sandler-CVPR2018-Mobilenetv2}
M.~Sandler, A.~Howard, M.~Zhu, A.~Zhmoginov, and L.-C. Chen.
\newblock Mobilenetv2: Inverted residuals and linear bottlenecks.
\newblock In {\em CVPR}, pp. 4510--4520, 2018.

\bibitem{Simonyan-ArXiv2014-VGG}
K.~Simonyan and A.~Zisserman.
\newblock Very deep convolutional networks for large-scale image recognition.
\newblock {\em arXiv preprint arXiv:1409.1556}, 2014.

\bibitem{Song-2022ISMAR-VTONShoes}
W.~Song, Y.~Gong, and Y.~Wang.
\newblock Vtonshoes: Virtual try-on of shoes in augmented reality on a mobile
  device.
\newblock In {\em ISMAR}, pp. 234--242, 2022.

\bibitem{Sun-IJCV2014-Quantitative}
D.~Sun, S.~Roth, and M.~J. Black.
\newblock A quantitative analysis of current practices in optical flow
  estimation and the principles behind them.
\newblock {\em IJCV}, pp. 115--137, 2014.

\bibitem{Wang-ECCV2018-Toward}
B.~Wang, H.~Zheng, X.~Liang, Y.~Chen, L.~Lin, and M.~Yang.
\newblock Toward characteristic-preserving image-based virtual try-on network.
\newblock In {\em ECCV}, pp. 589--604, 2018.

\bibitem{Wang-CVPR2023-YOLOv7}
C.-Y. Wang, A.~Bochkovskiy, and H.-Y.~M. Liao.
\newblock Yolov7: Trainable bag-of-freebies sets new state-of-the-art for
  real-time object detectors.
\newblock In {\em CVPR}, pp. 7464--7475, 2023.

\bibitem{Yang-CVPR2020-Towards}
H.~Yang, R.~Zhang, X.~Guo, W.~Liu, W.~Zuo, and P.~Luo.
\newblock Towards photo-realistic virtual try-on by adaptively
  generating-preserving image content.
\newblock In {\em CVPR}, pp. 7850--7859, 2020.

\bibitem{Zhang-CVPR2018-LPIPS}
R.~Zhang, P.~Isola, A.~A. Efros, E.~Shechtman, and O.~Wang.
\newblock The unreasonable effectiveness of deep features as a perceptual
  metric.
\newblock In {\em CVPR}, pp. 586--595, 2018.

\bibitem{Zhong-ACMMM2021-Mvton}
X.~Zhong, Z.~Wu, T.~Tan, G.~Lin, and Q.~Wu.
\newblock Mv-ton: Memory-based video virtual try-on network.
\newblock In {\em ACM Multimedia}, pp. 908--916, 2021.

\end{thebibliography}
\end{document}